%% file: main.tex
\PassOptionsToPackage{svgnames}{xcolor}
\AddToHook{package/unicode-math/before}{\RequirePackage{mathtools}}
\documentclass{allpurpose}

\input{include/supp_packages}

\title[Agentic Critical Training]{Agentic Critical Training}
\date{}
\paperlink[globe]{Project Page}{https://attention-is-all-i-need.github.io/ACT/}
\correspondence{\href{mailto:weizeliu@umd.edu}{\nolinkurl{weizeliu@umd.edu}};\allowbreak\space
  \href{mailto:furongh@umd.edu}{\nolinkurl{furongh@umd.edu}}}
\author{%
  \mbox{Weize Liu\affmark{1}}, \mbox{Minghui Liu\affmark{1}}, \mbox{Sy-Tuyen Ho\affmark{1}},
  \mbox{\apauthor{Yongkyun Lee}}, \mbox{\apauthor{Andrew Adams Schoen}},
  \mbox{Souradip Chakraborty\affmark{1}}, \mbox{Xiyao Wang\affmark{1,\ddag}},
  \mbox{Furong Huang\affmark{1,\forall,\ddag}}}
\affiliations{\affmark{1}University of Maryland, College Park\qquad\apaffiliation\\
  \affmark{\ddag}Equal advising}
\input{sections/s0_abs}
\hypersetup{pdfauthor={Weize Liu, Minghui Liu, Sy-Tuyen Ho, Yongkyun Lee, Andrew Adams Schoen, Souradip Chakraborty, Xiyao Wang, Furong Huang}}

\begin{document}
\maketitle
\begingroup
\renewcommand{\thefootnote}{}
\footnotetext{\textit{Corresponding author(s): Furong Huang; Email \href{mailto:furongh@umd.edu}{furongh@umd.edu}}}
\endgroup

\input{sections/s1_intro}

\input{sections/s2_method}

\input{sections/s3_related}

\input{sections/s4_exp}

\input{sections/s5_conc}

\section*{Acknowledgements}

Liu, Liu, Ho, Chakraborty, Wang, and Huang are supported by DARPA HR001124S0029-AIQ-FP-019, National Science Foundation TRAILS Institute (2229885). Private support was provided by Open Philanthropy and Apple. The Authors acknowledge the National Artificial Intelligence Research Resource (NAIRR) Pilot for contributing to this research result.

\clearpage
\bibliography{main}

\newpage
\appendix

\onecolumn
\input{sections/a0_details}
\clearpage
\input{sections/a1_agentic_case}
\clearpage
\input{sections/a2_general_case}

\end{document}

%% file: include/supp_packages.tex
\usepackage[all]{hypcap}
\setcitestyle{authoryear,round,comma}
\usepackage{multirow}
\usepackage{pgfplots}
\pgfplotsset{compat=1.18}
\usetikzlibrary{arrows.meta,calc,positioning}
\usepackage{cleveref}
\let\STATE\State
\let\FOR\For
\let\ENDFOR\EndFor
\let\IF\If
\let\ENDIF\EndIf
\AtBeginDocument{%
  \let\actmathcheckmark\checkmark
  \renewcommand{\checkmark}{\ensuremath{\actmathcheckmark}}%
  \let\actmathvdots\vdots
  \renewcommand{\vdots}{\ensuremath{\actmathvdots}}}
\crefname{table}{Table}{Tables}
\Crefname{table}{Table}{Tables}
\crefname{section}{Section}{Sections}
\Crefname{section}{Section}{Sections}
\definecolor{actblue}{HTML}{2F6B8A}
\definecolor{resultgray}{HTML}{8C8C8C}
\definecolor{resultlightgray}{HTML}{D9D9D9}
\definecolor{resultred}{HTML}{B85C5C}
\definecolor{resultorange}{HTML}{C58B3A}
\definecolor{resultgreen}{HTML}{4E8061}

%% file: sections/s0_abs.tex
\abstracttext{
Imitation learning (IL) teaches language-model agents to reproduce expert actions but not to distinguish them from plausible mistakes. Self-reflection methods expose models to alternatives yet use supervised fine-tuning (SFT) to imitate fixed rationales and actions. We introduce \textbf{Agentic Critical Training} (ACT), which uses reinforcement learning with verifiable rewards (RLVR) to train models to judge actions directly. At each expert-trajectory state, ACT pairs an expert action with an alternative sampled from the initial policy and randomizes their order. The model generates its own reasoning but is rewarded only for selecting the expert action. ACT reuses demonstrations, requires no reference rationales, and allows pair reuse across model sizes.
ACT is a warm-up before IL, optionally followed by RL; inference requires no candidate comparison. Across Qwen3-8B and Olmo-3-7B-Instruct on ALFWorld-ID, WebShop, and ScienceWorld, ACT yields average gains of 5.85 points over IL and 4.12 points over IL$\to$RL without ACT, while also improving ALFWorld-OOD. With Olmo on ScienceWorld, the full pipeline gains 15.36 points over CoT prompting and 9.23 points over IL$\to$RL without ACT. Both ACT$\to$IL and the full pipeline outperform supervised reflection baselines. Controls with fixed pairs or matched training durations show that the gains stem from the ACT objective rather than additional data or training. Without reasoning-specific post-training, the standalone ACT checkpoint achieves the highest mean among evaluated models on MATH-500 and GPQA-Diamond, showing that action comparison complements generation.
}

%% file: sections/s1_intro.tex
\section{Introduction}
\label{sec:intro}

Language-model agents make a sequence of choices while navigating websites \citep{yao2022webshop}, manipulating household environments \citep{shridharalfworld}, conducting scientific experiments \citep{wang2022scienceworld}, and calling tools \citep{patil2024gorilla}. A single plausible but incorrect action can derail the rest of a trajectory. Most agent post-training methods teach the model to generate the demonstrated next action, but do not explicitly teach it to compare that action with plausible alternatives. We study whether adding such a comparison objective can improve agent training.

\begin{figure}[t]
    \centering
    \includegraphics[width=0.96\textwidth]{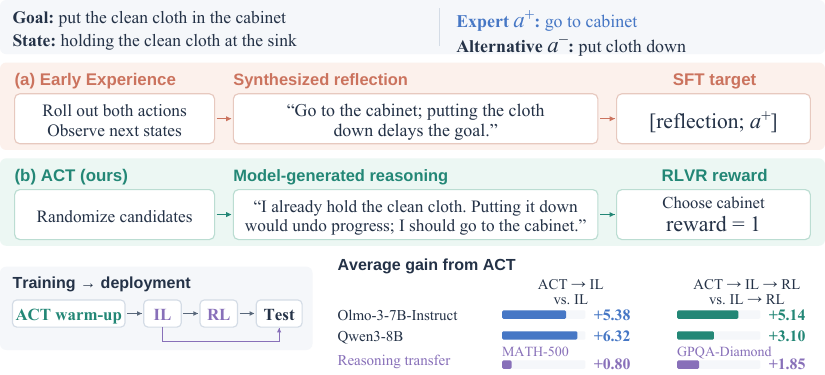}
    \caption{
        \textbf{ACT trains action judgment with a verifiable selection reward.}
        \textbf{(a)} Early Experience rolls out both candidates and uses SFT
        to imitate a synthesized reflection and expert action.
        \textbf{(b)} ACT randomizes the pair and rewards expert-action
        selection without a reference rationale.
        ACT precedes IL; RL continues from IL along the upper path,
        and the bypass goes directly to test.
        Blue and green bars show mean gains from adding ACT on
        ALFWorld-ID, WebShop, and ScienceWorld.
        Purple bars compare the ALFWorld-only Qwen3-8B ACT-stage checkpoint
        with the original Qwen3-8B model on general reasoning.
        All bars share a 0--7 point scale.
    }
    \label{fig:concept_overview}
\end{figure}

This limitation is clearest in imitation learning (IL). IL maximizes the likelihood of expert actions, so each training example identifies what the agent should do at a given state. It does not show which competing actions should be rejected or what makes them worse \citep{ross2011reduction,hussein2017imitation,pomerleau1991efficient}. Early Experience \citep{zhang2025agent} adds information about alternatives by rolling out both the expert and alternative actions, observing their resulting next states, and synthesizing a rationale that explains why the expert action is better. Although its targets contain more information than standard behavioral cloning, training still uses next-token prediction to reproduce a fixed rationale and the expert action. The model is given both the explanation and the answer, rather than trained to arrive at the judgment itself, as illustrated in~\Cref{fig:concept_overview}(a).

We propose \textbf{Agentic Critical Training} (ACT), shown in~\Cref{fig:concept_overview}(b), to train this judgment directly. For each state-action pair in an expert trajectory, we sample an alternative action from the initial policy and pair it with the expert action. The two candidates are shown in random order, and the model is asked to reason about them and select the better one. We train the model with reinforcement learning (RL), using a reward based on whether it selects the expert action. The reward does not evaluate the chain-of-thought \citep[CoT;][]{wei2022chain} or require the model to match a reference rationale. The supervision therefore evaluates the decision, while allowing the model to develop its own reasoning for that decision.

Building these pairs raises several practical issues. (1) The alternatives should resemble errors that the model is likely to make, (2) labels should be inexpensive to obtain, and (3) the input format should not provide shortcuts to the correct answer. We sample alternatives from the initial policy so that they reflect its own behavior, discard samples that match the expert action, and use the expert action as the preferred candidate under the assumption that expert actions are generally better. This construction requires no rationale generation. Randomizing the candidate order prevents the model from relying on a fixed position. We use one sampled alternative per state in our experiments and also find that pairs collected from an 8B model can train a 4B model, allowing the collection cost to be shared across model sizes.

\textbf{ACT is a training stage, not an inference-time reranking procedure.} It serves as a warm-up before IL, with optional RL continuing from the resulting IL checkpoint. We use \textbf{ACT stage} for the action-judgment warm-up, \textbf{full ACT pipeline} for ACT$\to$IL$\to$RL, and IL$\to$RL w/o ACT for the corresponding baseline. In this sense, ACT bridges the pretrained model and downstream policy training by first building the capacity to evaluate actions. At test time, the resulting policy generates actions directly without candidate pairs.

Across Qwen3-8B and Olmo-3-7B-Instruct~\citep{olmo2025olmo3}, ACT warm-up improves both IL and IL$\to$RL on ALFWorld, WebShop, and ScienceWorld, including ALFWorld-OOD. ACT$\to$IL and the full ACT pipeline also outperform supervised reflection baselines. Because these methods use the same expert and alternative actions, the comparison suggests that rewarding a correct judgment is more effective than reproducing a fixed rationale. Same-action-pair and matched-continuation controls show that ACT's gains are not explained by the data or additional training alone (\cref{tab:training_controls}).

The learned action judgment also transfers beyond the main setting. ACT pairs collected from Qwen3-8B effectively train Qwen3-4B, and the standalone ACT-stage checkpoint achieves the highest reported mean on MATH-500 and GPQA-Diamond without mathematical or scientific reasoning data during post-training. Together, these results suggest that expert trajectories can provide more than actions to imitate: paired with model-generated alternatives, they can also teach the model to distinguish better decisions from worse ones.

In summary, our contributions are:
\begin{enumerate}[leftmargin=*]
\item \textbf{Training objective.} We propose ACT, which uses RL to train an agent to select the better of an expert action and a model-generated alternative. The reward evaluates the selected action without requiring a reference rationale.
\item \textbf{Data construction.} We build contrastive pairs by sampling alternatives from the initial policy, filtering exact matches with the expert action, and randomizing candidate order, without generating reference rationales.
\item \textbf{Agent performance.} ACT$\to$IL and the full ACT pipeline improve over their no-ACT counterparts across three agent benchmarks and two model families, while outperforming supervised reflection baselines. Same-pair and matched-continuation controls isolate the benefit of the ACT objective.
\item \textbf{Generalization and analysis.} ACT warm-up improves agentic OOD performance and transfers across model sizes, while the ACT-stage checkpoint improves general reasoning without reasoning-specific post-training data. Causal interventions connect reasoning to action selection.
\end{enumerate}

%% file: sections/s2_method.tex
\section{Agentic Critical Training}
\label{sec:method}

\begin{figure*}[!htb]
    \centering
    \includegraphics[width=0.96\textwidth]{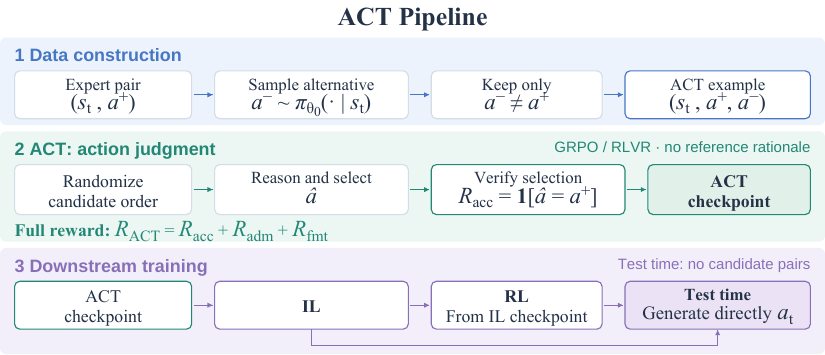}
    \caption{
        Overview of the ACT training pipeline.
        \textbf{Stage~1} constructs expert--alternative pairs
        from existing demonstrations.
        \textbf{Stage~2} trains action judgment with GRPO using
        the action-selection reward in \cref{eq:total_reward}.
        Its accuracy component verifies expert-action selection;
        admissibility and format terms are detailed in
        \cref{app:training_details}
        ($R_{\mathrm{adm}}$ is disabled for WebShop).
        \textbf{Stage~3} applies IL from the ACT checkpoint,
        followed by RL from the resulting IL checkpoint
        or the bypass to test.
        Both paths produce a policy that generates actions directly
        without candidate pairs at test time.
    }
    \label{fig:method_overview}
\end{figure*}

We present Agentic Critical Training (ACT), our approach to training LLM agents beyond pure imitation. We first describe the problem setting (\cref{sec:problem}), then detail ACT data construction (\cref{sec:data_collection}), and finally present the training pipeline (\cref{sec:training}). \Cref{fig:method_overview} provides an overview.

\subsection{Problem formulation}
\label{sec:problem}

We consider an agent operating in a sequential decision-making environment, formalized as a partially observable Markov decision process (POMDP) $\mathcal{M} = (\mathcal{S}, \mathcal{A}, \mathcal{T}, \Omega, \mathcal{O}, \mathcal{R}, \gamma)$, where $\mathcal{S}$ is the state space, $\mathcal{A}$ is the action space, $\mathcal{T}: \mathcal{S} \times \mathcal{A} \rightarrow \Delta(\mathcal{S})$ is the transition function, $\Omega$ is the observation space, $\mathcal{O}: \mathcal{S} \rightarrow \Delta(\Omega)$ is the observation function, $\mathcal{R}: \mathcal{S} \times \mathcal{A} \rightarrow \mathbb{R}$ is the reward function, and $\gamma \in [0,1)$ is the discount factor. In practice, the agent conditions on a textual context constructed from the task description, a truncated window of the most recent $k$ observation-action pairs, and the current observation (see prompt templates in \cref{app:prompt_templates}). We denote this observation-derived context as $s_t$ and use $s$ as shorthand throughout; the policy is thus $\pi_\theta(a_t|s_t)$.

Given an expert demonstration dataset $\mathcal{D}_{\text{expert}} = \left\{\tau^{(n)} = (s_1^{(n)}, a_1^{(n)}, \ldots, s_{T_n}^{(n)}, a_{T_n}^{(n)})\right\}_{n=1}^{N}$, the standard approach is imitation learning (IL), which maximizes the likelihood of expert actions:
\begin{equation}
\mathcal{L}_{\text{IL}}(\theta) = -\mathbb{E}_{(s,a)\sim\mathcal{D}_{\text{expert}}}\left[\log \pi_{\theta}(a|s)\right],
\label{eq:il_loss}
\end{equation}
where $\pi_{\theta}$ is the LLM policy. While effective at teaching agents to replicate expert behavior, IL provides no signal about the relative quality of different actions: the agent learns that $a_i$ is correct at context $s_i$, but not \emph{why} it is preferable to alternatives. Our goal is to train agents that develop this understanding.

\subsection{Data construction}
\label{sec:data_collection}

The core idea of ACT is to transform the learning objective from ``imitate the expert action'' to ``identify the better action,'' requiring the model to develop discriminative understanding of action quality. For each expert state-action pair $(s_i, a_i) \in \mathcal{D}_{\text{expert}}$, we construct an ACT example as follows:
\begin{enumerate}[leftmargin=*]
    \item \textbf{Sample alternatives.} Draw $K$ candidate actions $\{a_i^1, a_i^2, \ldots, a_i^K\}$ from an initial policy $\pi_{\theta_0}(\cdot|s_i)$.
    \item \textbf{Filter duplicates.} Remove candidates identical to the expert action: $\mathcal{A}_i^{\text{neg}} = \{a_i^j : a_i^j \neq a_i, j \in [K]\}$.
    \item \textbf{Construct pairs.} Pair the expert action $a_i^+ = a_i$ with each alternative $a_i^- \in \mathcal{A}_i^{\text{neg}}$ to form $|\mathcal{A}_i^{\text{neg}}|$ contrastive examples.
\end{enumerate}
This yields the ACT dataset $\mathcal{D}_{\text{critic}} = \{(s_i, a_i^+, a_i^-)\}_{i=1}^{M}$, where each example contains one expert action and one sampled alternative action. The key assumption is that the initial policy $\pi_{\theta_0}$ generates actions that are, on average, inferior to expert actions. This assumption is well-grounded in our setting: the benchmarks use constrained, discrete action spaces where expert trajectories are collected from specialized agents or human demonstrations, whereas the initial policy is a general-purpose pretrained LLM without task-specific training. We empirically find that the majority of sampled alternatives differ from the expert action and are identifiably suboptimal given the task context, which is consistent with the large performance gap between the initial policy and the expert.

\subsection{Training pipeline}
\label{sec:training}

After constructing $\mathcal{D}_{\text{critic}}$, the model first undergoes ACT with Group Relative Policy Optimization \citep[GRPO;][]{shao2024deepseekmath}, followed by IL with supervised fine-tuning. Optional RL then continues from the IL or ACT$\to$IL checkpoint. ACT uses an action-level selection reward; details of the ACT objective are provided in \cref{app:training_details}.

\paragraph{Agentic Critical Training.} The model is first trained on $\mathcal{D}_{\text{critic}}$ to identify the better action among two candidates. The ACT prompt is:

\begin{tcolorbox}[
  enhanced,
  boxrule=0.5pt,
  colback=AliceBlue,
  colframe=Navy,
  colbacktitle=Navy,
  fonttitle=\bfseries\color{white},
  title=ACT Prompt,
  arc=2mm,
  attach boxed title to top left={yshift=-2.5mm, xshift=3mm},
  boxed title style={arc=1mm},
  boxsep=5pt,
  left=5pt,
  right=5pt,
  top=4mm,
]
\small\texttt{%
Task: \{task\_description\}\\[2ex]
Below are the most recent \{k\} observations and actions:\\
\{history\}\\[2ex]
You are now at step \{current\_step\} and your current observation is:\\
$o_t$\\[2ex]
Your admissible actions of the current situation are:\\
$\mathcal{A}_{\text{admissible}}$\\[2ex]
Candidate Actions for the NEXT step:\\
Action 1: $a_{\sigma(1)}$\\
Action 2: $a_{\sigma(2)}$\\[2ex]
Which action is better as the NEXT step? Think about why, then output the chosen action inside <action>...</action> tags.
}
\end{tcolorbox}

Here $\sigma$ is a random permutation, so the expert action appears in either position with equal probability. 

Given this prompt, GRPO samples $G$ responses, each containing the model's reasoning and action selection. ACT provides no reference rationale: its verifiable reward depends on the final action selection rather than on matching reasoning content. This differs from the self-reflection variant of Early Experience~\citep{zhang2025agent}, which uses environment outcomes to synthesize rationales and then directly supervises the model to reproduce those rationales together with expert actions. ACT instead optimizes reasoning indirectly according to whether it supports correct selection. The resulting action-evaluation capability can then benefit direct action generation at test time. We examine whether the reasoning is behaviorally connected to selection through interventions in \cref{sec:rationale_intervention}.

\paragraph{IL and RL.} After ACT, the model first undergoes IL for direct action generation by maximizing the likelihood of expert actions. RL is optional and starts only from the IL or ACT$\to$IL checkpoint. Importantly, ACT is purely a training-time warm-up: the final policy generates actions directly without the contrastive selection format at test time.

\paragraph{ACT Reward Design.} The ACT stage uses the following composite action-selection reward:
\begin{equation}
R_{\text{ACT}}(s, y) = R_{\text{acc}}(a, a^+) + R_{\text{adm}}(a, \mathcal{A}_{\text{admissible}}) + R_{\text{fmt}}(y),
\label{eq:total_reward}
\end{equation}
where $y$ denotes the full ACT response and $a = \mathrm{extract}(y)$ denotes the selected action extracted from the \texttt{<action>...</action>} tags. We set $R_{\text{acc}} = 1$ if $a$ exactly matches the expert action $a^+$. We set $R_{\text{adm}} = 0.1$ if $a$ is admissible but does not match $a^+$, which provides partial credit for valid actions. We set $R_{\text{fmt}} = -0.5$ if the response lacks proper \texttt{<action>} tags. This action-level reward is used only for the ACT stage. Detailed definitions are provided in \cref{app:training_details}.

%% file: sections/s3_related.tex
\section{Related work}
\label{sec:related}

\paragraph{LLM-based Agents.}
LLM agents combine language reasoning with sequential action for web navigation \citep{yao2022webshop}, tool use \citep{schick2023toolformer,qin2024toolllm,su2025failure}, and multi-step tasks \citep{yao2022react,wei2022chain}. ReAct \citep{yao2022react} interleaves reasoning and acting, while Reflexion \citep{shinn2023reflexion} introduces verbal reflection at inference time. ACT instead trains action evaluation through RL, without relying on inference-time reflection prompts.

\paragraph{Training LLM Agents.}
Agent training commonly relies on imitation learning from expert demonstrations \citep{chen2023fireact,zeng2024agenttuning}. Early Experience \citep{zhang2025agent} augments demonstrations with outcomes from alternative actions, either predicting next states or synthesizing reflection targets; we compare with its self-reflection variant. SAND \citep{xia2025sand} synthesizes deliberation by critiquing alternatives, while Agent-R \citep{yuan2025agent} constructs revision trajectories from erroneous and corrected paths. These reflection and revision methods incorporate information beyond successful demonstrations, but still train the model to imitate fixed, constructed text targets. ACT instead uses RL to discriminate between expert and model-generated actions: the reward evaluates the final selection without prescribing a rationale.

\paragraph{Critique RL Training.}
RL has also been used to train critics for reward modeling \citep{zhang2025r1,chen2025rm} or to improve policies through critique generation \citep{wang2025llava,ruan2025critique}. These methods primarily critique standalone outputs in single-turn settings. ACT instead learns to distinguish expert from suboptimal actions within a multi-turn decision process.

\paragraph{Agentic RL.}
Agentic RL extends post-training beyond chat alignment \citep{ouyang2022training,rafailov2023direct} to multi-turn, long-horizon interaction \citep{zhanglandscape}. RL with verifiable rewards can elicit reasoning without supervised chains of thought \citep{guo2025deepseek}; GRPO \citep{shao2024deepseekmath}, GiGPO \citep{fenggroup}, and Search-R1 \citep{jinsearch} develop this idea for group-relative optimization, agent credit assignment, and interactive search. ACT contributes a complementary warm-up objective: learning to discriminate expert from suboptimal actions before policy training.

%% file: sections/s4_exp.tex
\section{Experiments}
\label{sec:experiments}

We evaluate our approach on three diverse benchmarks spanning different agent capabilities. Unless otherwise noted, reported values are the mean $\pm$ standard deviation across three runs.

\subsection{Experimental setup}

\paragraph{Benchmarks.}
We evaluate on \textbf{ALFWorld}~\citep{shridharalfworld} for embodied household tasks, \textbf{WebShop}~\citep{yao2022webshop} for web shopping, and \textbf{ScienceWorld}~\citep{wang2022scienceworld} for multi-step scientific experiments. We report ALFWorld's seen and unseen splits as in-distribution (ID) and out-of-distribution (OOD), respectively. Full evaluation protocols are provided in \cref{app:revision_protocol}.

\paragraph{Training methods.}
All training-based methods use the same expert trajectories $\mathcal{D}_{\text{expert}}$. We compare prompting, the standalone \textbf{ACT stage}, \textbf{IL}, \textbf{ACT$\to$IL}, \textbf{IL$\to$RL w/o ACT}, and the \textbf{full ACT pipeline} (ACT$\to$IL$\to$RL). Here IL uses next-token supervision on expert actions, whereas RL starts from an IL checkpoint and uses GRPO. Details, hyperparameters, and prompts are provided in \cref{app:experiments}.

\subsection{Main results}

\Cref{tab:main_results,tab:olmo_main_results} present our main results on Qwen3-8B~\citep{yang2025qwen3} and Olmo-3-7B-Instruct~\citep{olmo2025olmo3}, respectively. The two tables use the same methods and evaluation settings across all three benchmarks.

\paragraph{ACT warm-up improves downstream performance across model families.} The standalone ACT-stage checkpoint improves over prompting baselines but does not match IL or IL$\to$RL w/o ACT in absolute agent performance, as it trains the model to \emph{judge} actions rather than generate them directly. When used as a warm-up stage, adding ACT improves both downstream pipelines, and the full ACT pipeline achieves the highest mean in all four evaluation settings on both Qwen3-8B and Olmo-3-7B-Instruct. On Qwen3, adding ACT improves IL by 6.32 points and IL$\to$RL w/o ACT by 3.10 points on average across ALFWorld-ID, WebShop, and ScienceWorld. On Olmo 3, the corresponding average gains are 5.38 and 5.14 points. The Olmo ScienceWorld gain is particularly large: the full ACT pipeline improves the mean score by 15.36 points over CoT prompting and by 9.23 points over IL$\to$RL w/o ACT.

\paragraph{ACT$\to$IL and the full ACT pipeline outperform supervised reflection baselines.} We compare with Early Experience (Self-Reflection)~\citep{zhang2025agent}, SAND~\citep{xia2025sand}, and Agent-R~\citep{yuan2025agent}, which augment action supervision by training models to generate additional self-reflection, deliberation, or revision text. Across both model families, ACT$\to$IL and the full ACT pipeline outperform all three baselines in every evaluation setting. This suggests that optimizing action-quality reasoning through outcome-based RL is more effective than learning additional synthesized reasoning text through supervised training. Baseline implementations appear in \cref{app:baseline_details}.

\paragraph{ACT warm-up improves OOD generalization.} On ALFWorld-OOD, adding ACT before IL improves performance from 79.85 to 83.58 on Qwen3 and from 56.97 to 67.41 on Olmo 3. The full ACT pipeline also improves over IL$\to$RL w/o ACT from 90.80 to 91.54 on Qwen3 and from 77.11 to 83.83 on Olmo 3. Thus, the action judgment acquired during the ACT stage transfers to unseen task configurations on both backbones.

\begin{table*}[t]
\centering
\caption{Main results on Qwen3-8B. ALFWorld and WebShop report success rates (\%); ScienceWorld reports mean episode score (0--100).}
\label{tab:main_results}
\begin{tabular}{@{\hspace{3pt}}l*{4}{c}@{\hspace{3pt}}}
\toprule
\multirow{2}{*}{\textbf{Method}} & \multicolumn{2}{c}{\textbf{ALFWorld}} & \textbf{WebShop} & \textbf{ScienceWorld} \\
\cmidrule(r){2-3}
& ID & OOD & & \\
\midrule
Prompt w/o CoT thinking & 32.62$\pm$2.63 & 24.88$\pm$2.88 & 18.20$\pm$0.85 & 16.99$\pm$2.85 \\
Prompt w/ CoT thinking & 50.00$\pm$4.77 & 39.30$\pm$7.72 & 15.53$\pm$0.34 & 18.68$\pm$0.34 \\
\midrule
ACT stage & 62.14$\pm$7.72 & 59.45$\pm$9.23 & 21.53$\pm$4.58 & 30.04$\pm$2.13 \\
\midrule
Imitation Learning (IL) & 85.71$\pm$0.58 & 79.85$\pm$2.79 & 28.00$\pm$1.31 & 21.94$\pm$2.65 \\
Early Experience & 86.43$\pm$1.55 & 79.10$\pm$10.57 & 30.93$\pm$1.23 & 22.82$\pm$1.30 \\
SAND & 85.71$\pm$1.16 & 83.09$\pm$0.93 & 20.27$\pm$1.36 & 29.41$\pm$0.46 \\
Agent-R & 87.86$\pm$1.17 & 81.09$\pm$0.93 & 24.07$\pm$4.20 & 16.95$\pm$1.79 \\
ACT $\to$ IL & 89.76$\pm$1.88 & 83.58$\pm$3.80 & 32.00$\pm$1.18 & 32.85$\pm$0.15 \\
\midrule
IL $\to$ RL w/o ACT & 94.28$\pm$1.54 & 90.80$\pm$0.70 & 39.53$\pm$1.06 & 30.52$\pm$0.46 \\
Full ACT pipeline & \textbf{96.19$\pm$0.67} & \textbf{91.54$\pm$0.35} & \textbf{42.80$\pm$0.59} & \textbf{34.63$\pm$0.33} \\
\bottomrule
\end{tabular}
\end{table*}

\begin{table*}[t]
\centering
\caption{Main results on Olmo-3-7B-Instruct. ALFWorld and WebShop report success rates (\%); ScienceWorld reports mean episode score (0--100).}
\label{tab:olmo_main_results}
\begin{tabular}{@{\hspace{3pt}}l*{4}{c}@{\hspace{3pt}}}
\toprule
\multirow{2}{*}{\textbf{Method}} & \multicolumn{2}{c}{\textbf{ALFWorld}} & \textbf{WebShop} & \textbf{ScienceWorld} \\
\cmidrule(r){2-3}
& ID & OOD & & \\
\midrule
Prompt w/o CoT thinking & 0.71$\pm$0.58 & 0.25$\pm$0.35 & 11.47$\pm$0.34 & 3.47$\pm$2.04 \\
Prompt w/ CoT thinking & 11.91$\pm$0.67 & 10.94$\pm$2.14 & 12.47$\pm$0.47 & 6.31$\pm$3.97 \\
\midrule
ACT stage & 16.90$\pm$1.69 & 11.94$\pm$3.71 & 11.53$\pm$1.31 & 12.03$\pm$0.92 \\
\midrule
Imitation Learning (IL) & 62.62$\pm$3.37 & 56.97$\pm$9.33 & 12.87$\pm$9.02 & 8.10$\pm$0.19 \\
Early Experience & 42.62$\pm$3.51 & 35.32$\pm$6.40 & 13.27$\pm$7.04 & 8.97$\pm$0.71 \\
SAND & 36.91$\pm$1.78 & 28.11$\pm$3.52 & 7.00$\pm$0.33 & 8.11$\pm$0.31 \\
Agent-R & 66.43$\pm$4.67 & 63.93$\pm$4.93 & 13.40$\pm$0.86 & 9.25$\pm$0.70 \\
ACT $\to$ IL & 74.52$\pm$2.21 & 67.41$\pm$4.06 & 14.80$\pm$7.27 & 10.40$\pm$0.89 \\
\midrule
IL $\to$ RL w/o ACT & 81.67$\pm$2.21 & 77.11$\pm$5.91 & 23.40$\pm$1.23 & 12.44$\pm$1.54 \\
Full ACT pipeline & \textbf{87.14$\pm$3.08} & \textbf{83.83$\pm$0.35} & \textbf{24.13$\pm$0.41} & \textbf{21.67$\pm$0.61} \\
\bottomrule
\end{tabular}
\end{table*}

\subsection{Training-objective and continuation controls}
\label{sec:training_controls}

\begin{table*}[t]
\centering
\caption{Training-objective and continuation controls on Qwen3-8B. ALFWorld and WebShop report success rates (\%); ScienceWorld reports mean episode score (0--100). Objective controls use the same action pairs and downstream IL$\to$RL pipeline. Continuation comparisons match six total epochs for IL$\to$IL versus ACT$\to$IL and nine total epochs for IL$\to$RL$\to$RL versus the full ACT pipeline. Bold denotes the highest mean within each block.}
\label{tab:training_controls}
\begin{tabular}{@{\hspace{3pt}}l*{4}{c}@{\hspace{3pt}}}
\toprule
\multirow{2}{*}{\textbf{Method}} & \multicolumn{2}{c}{\textbf{ALFWorld}} & \textbf{WebShop} & \textbf{ScienceWorld} \\
\cmidrule(r){2-3}
& ID & OOD & & \\
\midrule
\multicolumn{5}{c}{\textbf{(a) Matched IL continuation}} \\
Imitation Learning (IL) & 85.71$\pm$0.58 & 79.85$\pm$2.79 & 28.00$\pm$1.31 & 21.94$\pm$2.65 \\
IL $\to$ IL & 84.76$\pm$0.89 & 79.11$\pm$3.71 & 31.40$\pm$2.57 & 22.01$\pm$2.01 \\
ACT $\to$ IL & \textbf{89.76$\pm$1.88} & \textbf{83.58$\pm$3.80} & \textbf{32.00$\pm$1.18} & \textbf{32.85$\pm$0.15} \\
\midrule
\multicolumn{5}{c}{\textbf{(b) ACT-stage objective and matched RL controls}} \\
IL $\to$ RL w/o ACT & 94.28$\pm$1.54 & 90.80$\pm$0.70 & 39.53$\pm$1.06 & 30.52$\pm$0.46 \\
IL $\to$ RL $\to$ RL & 90.00$\pm$2.10 & 89.80$\pm$3.47 & 33.40$\pm$3.82 & 20.14$\pm$1.09 \\
Full pipeline (SFT-on-choice) & 94.76$\pm$1.34 & \textbf{92.78$\pm$1.41} & 37.73$\pm$1.48 & 30.10$\pm$0.19 \\
Full pipeline (DPO) & 93.57$\pm$0.58 & 90.05$\pm$1.53 & 41.40$\pm$0.99 & 33.56$\pm$0.27 \\
Full pipeline (no-reasoning) & 94.52$\pm$0.67 & 92.54$\pm$0.61 & 38.47$\pm$2.51 & 28.12$\pm$2.33 \\
Full ACT pipeline & \textbf{96.19$\pm$0.67} & 91.54$\pm$0.35 & \textbf{42.80$\pm$0.59} & \textbf{34.63$\pm$0.33} \\
\bottomrule
\end{tabular}
\end{table*}

We conduct two sets of controls on Qwen3-8B. First, we vary how the ACT stage is trained while holding its contrastive pairs and downstream IL$\to$RL pipeline fixed. Second, we compare ACT warm-up with additional IL or RL training with matched budgets. Each stage lasts three epochs.

\paragraph{ACT-stage objective controls.} We replace ACT-stage RL with SFT-on-choice or DPO, or retain RL while removing the reasoning instruction. As shown in \cref{tab:training_controls}, the full ACT pipeline achieves the highest mean on ALFWorld-ID, WebShop, and ScienceWorld, while the SFT-on-choice pipeline leads on ALFWorld-OOD. The full ACT objective therefore provides the strongest overall performance across the four settings. Detailed control implementations are provided in \cref{app:baseline_details}.

\paragraph{Matched IL continuation.} We compare IL$\to$IL, where a three-epoch IL checkpoint receives three additional IL epochs, with ACT$\to$IL, where three ACT epochs are followed by three IL epochs. ACT$\to$IL exceeds IL$\to$IL by 5.00 points on ALFWorld-ID, 4.47 on ALFWorld-OOD, 0.60 on WebShop, and 10.84 on ScienceWorld. Extending IL barely changes the ScienceWorld score (21.94 to 22.01), whereas ACT$\to$IL reaches 32.85.

\paragraph{Matched RL continuation.} We compare the full ACT pipeline (ACT$\to$IL$\to$RL) with IL$\to$RL$\to$RL, which adds three RL epochs to the standard IL$\to$RL checkpoint. Both pipelines receive nine total epochs. The full ACT pipeline exceeds IL$\to$RL$\to$RL by 6.19 points on ALFWorld-ID, 1.74 on ALFWorld-OOD, 9.40 on WebShop, and 14.49 on ScienceWorld.

\paragraph{Additional training alone is insufficient.} Three additional IL epochs lower ALFWorld-ID and OOD performance, while three additional RL epochs reduce the mean in all four settings. In contrast, ACT warm-up improves over the corresponding IL or IL$\to$RL baseline in every setting. These controls show that allocating the additional stage to action judgment is more effective than repeating the downstream action-generation objective.

\subsection{Causal rationale intervention}
\label{sec:rationale_intervention}

We evaluate whether the reasoning generated by the standalone ACT-stage checkpoint is behaviorally connected to its final action selection through inference-time interventions on 400 held-out ALFWorld action pairs~\citep{liu2026ica}. We compare unconstrained generation, direct selection without a rationale, and generation after injecting a plausible but incorrect rationale favoring the alternative action.

\begin{figure}[t]
\centering
\resizebox{0.90\textwidth}{!}{\input{figures/rationale_intervention}}
\caption{Causal rationale intervention on 400 held-out ALFWorld pairs. \textbf{(a)} Choice accuracy under free generation, direct selection without a rationale, and an injected incorrect rationale. \textbf{(b)} Fraction of examples in which the model follows the action favored by the injected incorrect rationale. Exact values are provided in \cref{tab:rationale_intervention}.}
\label{fig:rationale_intervention}
\end{figure}

As shown in \cref{fig:rationale_intervention}, the ACT checkpoint retains 50.75\% accuracy while being actively guided toward the incorrect action, compared with 2.25\% for the original model, and reduces the rate of following the injected action from 81.75\% to 28.25\%. Removing the rationale lowers its accuracy from 86.50\% to 80.75\%, showing that its generated reasoning is behaviorally relevant to the final choice. Full intervention details are provided in \cref{app:rationale_protocol}.

\paragraph{Case studies: action evaluation supports agentic decision-making.} On ALFWorld, an IL model repeats the same failed action until termination, while the model trained with the full ACT pipeline diagnoses the root cause and self-corrects. On WebShop, an IL trajectory proceeds to purchase without checking a violated price constraint, illustrating the need for state-aware action evaluation rather than rigid execution. Detailed traces are provided in \cref{app:agentic_case_studies}.

\subsection{Generalization to general reasoning benchmarks}
\label{sec:general_reasoning}

Beyond agentic tasks, we investigate whether the critical reasoning capabilities acquired through ACT transfer to general reasoning benchmarks. We directly evaluate the Qwen3-8B checkpoints trained on ALFWorld agentic data, including IL, IL$\to$RL w/o ACT, and the standalone ACT-stage checkpoint, on MATH-500 \citep{hendrycks2021measuring} and GPQA-Diamond \citep{rein2024gpqa}. None of these checkpoints receives mathematical or scientific reasoning data during our training.

\begin{figure}[t]
\centering
\resizebox{0.90\textwidth}{!}{\input{figures/general_reasoning}}
\caption{General-reasoning accuracy after training only on ALFWorld agent data. Bars and error bars report the mean $\pm$ standard deviation over three runs. Exact values are provided in \cref{tab:general_reasoning}.}
\label{fig:general_reasoning}
\end{figure}

As shown in \cref{fig:general_reasoning}, the training paradigms have different effects on general reasoning. IL and the supervised self-reflection baseline, both optimized with next-token objectives against fixed targets, do not yield consistent improvements from agentic training. On MATH-500, both maintain performance comparable to the original model. On GPQA-Diamond, IL degrades performance by 6.91 points compared with the CoT prompting baseline (44.61\% vs.\ 51.52\%), while the self-reflection baseline recovers to approximately the prompting baseline (51.85\%). Thus, enriching supervised targets with reflection text does not produce the cross-domain reasoning gains observed with ACT.

Among the reflection baselines, SAND notably degrades MATH-500 to 81.00$\pm$0.86\%, while Agent-R reaches 86.53$\pm$0.81\% on MATH-500 and 52.86$\pm$1.04\% on GPQA-Diamond. Neither method surpasses the CoT prompting baseline on both benchmarks simultaneously.

IL$\to$RL w/o ACT roughly preserves the original model's performance on both benchmarks. The standalone ACT-stage checkpoint achieves the highest mean among the models tested on both MATH-500 (87.73$\pm$0.19\%) and GPQA-Diamond (53.37$\pm$0.63\%) despite being trained exclusively on agentic data, while IL degrades GPQA-Diamond by 6.91 points. This result indicates that RL in agentic environments, when combined with the ACT objective, can serve as a viable pathway for enhancing general reasoning capabilities.

\paragraph{Case studies: self-verification and reasoning collapse.} We examine reasoning traces and observe two phenomena. First, the standalone ACT-stage checkpoint exhibits \emph{self-verification}: on a difficult GPQA-Diamond problem, the model checks its answer by substituting back into the original equations, systematically eliminating incorrect options. Second, an IL trace on another GPQA-Diamond problem is substantially longer and more meandering than the corresponding standalone ACT-stage trace. Detailed examples are provided in \cref{app:case_studies}.

%% file: figures/rationale_intervention.tex
\noindent
\begin{minipage}[t]{0.66\textwidth}
\centering
\begin{tikzpicture}
\begin{axis}[
    width=0.96\linewidth,
    height=5.2cm,
    ybar,
    bar width=15pt,
    ymin=0,
    ymax=105,
    ytick={0,25,50,75,100},
    ylabel={Choice accuracy (\%)},
    symbolic x coords={Free,No rationale,Incorrect rationale},
    xtick=data,
    enlarge x limits=0.13,
    xticklabel style={align=center,font=\small},
    tick label style={font=\small},
    label style={font=\small},
    title={\textbf{(a) Accuracy under rationale interventions}},
    title style={font=\small},
    ymajorgrids=true,
    grid style={draw=gray!25},
    axis x line*=bottom,
    axis y line*=left,
    axis line style={black},
    nodes near coords,
    nodes near coords style={font=\scriptsize},
    legend style={draw=none,font=\scriptsize,at={(1.02,1.01)},anchor=north east,
        inner sep=1pt},
    legend image code/.code={\draw[#1] (0cm,-0.07cm) rectangle (0.16cm,0.07cm);},
]
\addplot[fill=resultlightgray,draw=gray!70] coordinates {
    (Free,45.00) (No rationale,46.25) (Incorrect rationale,2.25)
};
\addlegendentry{Original Qwen3-8B}
\addplot[fill=actblue,draw=actblue!70!black] coordinates {
    (Free,86.50) (No rationale,80.75) (Incorrect rationale,50.75)
};
\addlegendentry{ACT checkpoint}
\end{axis}
\end{tikzpicture}
\end{minipage}\hfill
\begin{minipage}[t]{0.31\textwidth}
\centering
\begin{tikzpicture}
\begin{axis}[
    width=0.92\linewidth,
    height=5.2cm,
    ybar=0pt,
    bar width=24pt,
    ymin=0,
    ymax=105,
    ytick={0,25,50,75,100},
    ylabel={Follow rate (\%)},
    symbolic x coords={Original,ACT},
    xtick={Original,ACT},
    xticklabels={Original,ACT},
    enlarge x limits=0.32,
    xticklabel style={font=\small,align=center},
    tick label style={font=\small},
    label style={font=\small},
    title={\textbf{(b) Follows injected action}},
    title style={font=\small},
    ymajorgrids=true,
    grid style={draw=gray!25},
    axis x line*=bottom,
    axis y line*=left,
    axis line style={black},
    nodes near coords,
    nodes near coords style={font=\scriptsize},
]
\addplot[fill=resultlightgray,draw=gray!70,bar shift=0pt] coordinates {(Original,81.75)};
\addplot[fill=actblue,draw=actblue!70!black,bar shift=0pt] coordinates {(ACT,28.25)};
\end{axis}
\end{tikzpicture}
\end{minipage}

%% file: figures/general_reasoning.tex
\newcommand{\reasonpoint}[6]{%
    \draw[#6,line width=0.8pt] (axis cs:#2,#1)--(axis cs:#3,#1);
    \draw[#6,line width=0.8pt] ([yshift=-2pt]axis cs:#2,#1)--([yshift=2pt]axis cs:#2,#1);
    \draw[#6,line width=0.8pt] ([yshift=-2pt]axis cs:#3,#1)--([yshift=2pt]axis cs:#3,#1);
    \filldraw[fill=#6,draw=gray!70!black,line width=0.3pt] (axis cs:#4,#1) circle (2.2pt);
    \node[anchor=west,font=\scriptsize,xshift=2pt] at (axis cs:#3,#1) {#5};
}
\newcommand{\ilrlwoact}{IL \ensuremath{\to} RL w/o ACT}
\definecolor{sandcolor}{HTML}{9B6B3D}
\definecolor{agentrcolor}{HTML}{7B4FA0}
\noindent
\begin{minipage}[t]{0.49\textwidth}
\centering
\begin{tikzpicture}
\begin{axis}[
    width=0.78\linewidth,
    height=7.0cm,
    xmin=77.5,
    xmax=90,
    xtick={78,82,86,89},
    ytick={1,2,3,4,5,6,7,8},
    yticklabels={ACT stage,\ilrlwoact,Agent-R,SAND,Early Experience,Imitation Learning,Prompt w/ CoT,Prompt w/o CoT},
    ymin=0.5,
    ymax=8.5,
    xlabel={Accuracy (\%)},
    title={\textbf{MATH-500}},
    title style={font=\small},
    tick label style={font=\small},
    label style={font=\small},
    xmajorgrids=true,
    grid style={draw=gray!25},
    axis x line*=bottom,
    axis y line*=left,
    clip=false,
]
\reasonpoint{8}{78.27}{78.93}{78.60}{78.60}{resultlightgray}
\reasonpoint{7}{86.19}{87.67}{86.93}{86.93}{resultgray}
\reasonpoint{6}{86.67}{87.33}{87.00}{87.00}{resultred}
\reasonpoint{5}{86.61}{87.11}{86.86}{86.86}{resultorange}
\reasonpoint{4}{80.14}{81.86}{81.00}{81.00}{sandcolor}
\reasonpoint{3}{85.72}{87.34}{86.53}{86.53}{agentrcolor}
\reasonpoint{2}{86.30}{87.84}{87.07}{87.07}{resultgreen}
\reasonpoint{1}{87.54}{87.92}{87.73}{87.73}{actblue}
\end{axis}
\end{tikzpicture}
\end{minipage}\hfill
\begin{minipage}[t]{0.49\textwidth}
\centering
\begin{tikzpicture}
\begin{axis}[
    width=0.94\linewidth,
    height=7.0cm,
    xmin=40,
    xmax=57,
    xtick={40,44,48,52,56},
    ytick={1,2,3,4,5,6,7,8},
    yticklabels={},
    ymin=0.5,
    ymax=8.5,
    xlabel={Accuracy (\%)},
    title={\textbf{GPQA-Diamond}},
    title style={font=\small},
    tick label style={font=\small},
    label style={font=\small},
    xmajorgrids=true,
    grid style={draw=gray!25},
    axis x line*=bottom,
    axis y line*=left,
    clip=false,
]
\reasonpoint{8}{41.84}{44.02}{42.93}{42.93}{resultlightgray}
\reasonpoint{7}{49.63}{53.41}{51.52}{51.52}{resultgray}
\reasonpoint{6}{43.66}{45.56}{44.61}{44.61}{resultred}
\reasonpoint{5}{51.22}{52.48}{51.85}{51.85}{resultorange}
\reasonpoint{4}{50.88}{54.18}{52.53}{52.53}{sandcolor}
\reasonpoint{3}{51.82}{53.90}{52.86}{52.86}{agentrcolor}
\reasonpoint{2}{51.04}{53.68}{52.36}{52.36}{resultgreen}
\reasonpoint{1}{52.74}{54.00}{53.37}{53.37}{actblue}
\end{axis}
\end{tikzpicture}
\end{minipage}
\let\reasonpoint\relax
\let\ilrlwoact\relax

%% file: sections/s5_conc.tex
\section{Conclusion}
\label{sec:conclusion}

We introduced ACT, which uses RL to train action judgment by contrasting expert and model-generated actions. Unlike supervised reflection methods, ACT evaluates the selected action without prescribing a reference rationale. ACT serves as a warm-up before IL, with optional RL continuing from the resulting IL checkpoint. Across Qwen3-8B and Olmo-3-7B-Instruct on three agent benchmarks, adding ACT improves both downstream pipelines; the full ACT pipeline achieves the highest mean in every evaluation setting and outperforms supervised reflection baselines. Same-action-pair and compute-matched controls show that these gains depend on how the additional stage is trained, rather than additional optimization alone. The improvements extend to unseen agent tasks, while the standalone ACT-stage checkpoint attains the highest mean on both MATH-500 and GPQA-Diamond despite receiving only agentic training. Together, these results suggest that learning to compare actions is a useful training objective whose benefits transfer to agent policies and general reasoning.

%% file: sections/a0_details.tex
\section{Experimental details}
\label{app:experiments}

\subsection{Training formulation and algorithm}
\label{app:training_details}

\paragraph{Reward Function Design.}
\label{sec:reward}
Our composite reward function consists of three components. Given a full generated response $y$ for state $s$, we first extract the action span inside the \texttt{<action>...</action>} tags and denote the extracted action by $a = \mathrm{extract}(y)$. The semantic rewards are applied to the extracted action $a$, while the format reward is applied to the full response $y$:
\begin{equation}
R_{\text{ACT}}(s, y) = R_{\text{acc}}(a, a^+) + R_{\text{adm}}(a, \mathcal{A}_{\text{admissible}}) + R_{\text{fmt}}(y).
\label{eq:total_reward_full}
\end{equation}
If the response does not contain a valid tagged action span, we set $a = \emptyset$, so the response receives zero semantic reward and only the format penalty applies.

The \emph{accuracy reward} measures exact match between the extracted action and the expert action:
\begin{equation}
R_{\text{acc}}(a, a^+) = \begin{cases}
1.0 & \text{if } \text{normalize}(a) = \text{normalize}(a^+) \\
0.0 & \text{otherwise}
\end{cases}
\label{eq:acc_reward}
\end{equation}

The \emph{admissible action reward} provides partial credit for extracted actions that are valid but suboptimal in environments with defined action spaces:
\begin{equation}
R_{\text{adm}}(a, \mathcal{A}_{\text{admissible}}) = \begin{cases}
0.1 & \text{if } a \neq a^+ \land a \in \mathcal{A}_{\text{admissible}} \\
0.0 & \text{otherwise}
\end{cases}
\label{eq:adm_reward}
\end{equation}

For WebShop, the admissible-action list contains the current page's clickable elements and, when a search bar is present, a \texttt{search[<query>]} template. This template represents arbitrary non-empty queries rather than enumerating concrete search actions. Since search queries are open-ended, we disable $R_{\text{adm}}$ for ACT on WebShop and use only $R_{\text{acc}}$ and $R_{\text{fmt}}$.

The \emph{format reward} penalizes full responses that lack proper action tags:
\begin{equation}
R_{\text{fmt}}(y) = \begin{cases}
0.0 & \text{if action tags present} \\
-0.5 & \text{otherwise}
\end{cases}
\label{eq:fmt_reward}
\end{equation}

\paragraph{GRPO Algorithm.} The ACT stage and optional downstream RL stage use Group Relative Policy Optimization (GRPO)~\citep{shao2024deepseekmath}, while the intervening IL stage uses supervised fine-tuning. GRPO is a variant of proximal policy optimization~\citep{schulman2017proximal} that eliminates the need for a learned value function by estimating advantages from group-level reward statistics. Given a prompt (state) $s$, GRPO samples a group of $G$ responses $\{y^{(1)}, y^{(2)}, \ldots, y^{(G)}\}$ from the current policy $\pi_\theta$. Each response receives the reward defined for the current training stage, and the advantage for each response is computed relative to the group statistics:
\begin{equation}
\hat{A}^{(g)} = \frac{r^{(g)} - \bar{r}}{\sigma_r + \epsilon},
\label{eq:grpo_advantage}
\end{equation}
where $\bar{r} = \frac{1}{G}\sum_{g=1}^{G} r^{(g)}$ is the group mean reward, $\sigma_r = \sqrt{\frac{1}{G}\sum_{g=1}^{G}(r^{(g)} - \bar{r})^2}$ is the group standard deviation, and $\epsilon$ is a small constant for numerical stability. The GRPO objective combines the policy gradient with KL regularization:
\begin{align}
\mathcal{L}_{\text{GRPO}}(\theta) &= -\mathbb{E}_{s \sim \mathcal{D}}\mathbb{E}_{y^{(g)} \sim \pi_\theta(\cdot|s)}\bigg[\min\Big(\rho^{(g)} \hat{A}^{(g)}, \nonumber\\
&\quad \text{clip}(\rho^{(g)}, 1-\epsilon_c, 1+\epsilon_c) \hat{A}^{(g)}\Big)\bigg] \nonumber\\
&\quad + \beta \cdot D_{\text{KL}}(\pi_\theta \| \pi_{\text{ref}}),
\label{eq:grpo_loss}
\end{align}
where $\rho^{(g)} = \frac{\pi_\theta(y^{(g)}|s)}{\pi_{\theta_{\text{old}}}(y^{(g)}|s)}$ is the importance sampling ratio, $\epsilon_c$ is the clipping threshold, $\beta$ is the KL penalty coefficient, and $\pi_{\text{ref}}$ is the reference policy. For compactness, the expression suppresses token indices; the clipped objective is applied tokenwise in implementation. We set $\beta=0$ in all experiments (\cref{tab:hyperparams}).

\paragraph{Training Algorithm.} \Cref{alg:critical_training} summarizes the complete ACT procedure. The algorithm consists of two phases: data collection, where we construct the ACT dataset by sampling alternatives from the initial policy, and GRPO training, where we optimize the policy using verifiable rewards based on action correctness.

\begin{algorithm}[tb]
\caption{ACT with GRPO}
\label{alg:critical_training}
\begin{algorithmic}
\STATE {\bfseries Input:} Expert dataset $\mathcal{D}_{\text{expert}}$, initial policy $\pi_{\theta_0}$, number of candidate samples $K$, group size $G$
\STATE {\bfseries Output:} Trained policy $\pi_{\theta}$
\STATE
\STATE // \textit{Phase 1: Data Collection}
\STATE $\mathcal{D}_{\text{critic}} \leftarrow \emptyset$
\FOR{each $(s_i, a_i^+) \in \mathcal{D}_{\text{expert}}$}
    \STATE Sample $\{a_i^1, \ldots, a_i^K\} \sim \pi_{\theta_0}(\cdot|s_i)$
    \STATE $\mathcal{A}_i^{\text{neg}} \leftarrow \{a_i^j : a_i^j \neq a_i^+\}$
    \IF{$|\mathcal{A}_i^{\text{neg}}| > 0$}
        \FOR{each $a_i^- \in \mathcal{A}_i^{\text{neg}}$}
            \STATE $\mathcal{D}_{\text{critic}} \leftarrow \mathcal{D}_{\text{critic}} \cup \{(s_i, a_i^+, a_i^-)\}$
        \ENDFOR
    \ENDIF
\ENDFOR
\STATE
\STATE // \textit{Phase 2: GRPO Training}
\STATE Initialize $\theta \leftarrow \theta_0$, $\pi_{\text{ref}} \leftarrow \pi_{\theta_0}$
\FOR{each training iteration}
    \STATE Set $\theta_{\text{old}} \leftarrow \theta$
    \STATE Sample batch $\mathcal{B} \subset \mathcal{D}_{\text{critic}}$
    \FOR{each $(s, a^+, a^-) \in \mathcal{B}$}
        \STATE Construct ACT prompt $p$ with randomized positions
        \STATE Sample $\{y^{(1)}, \ldots, y^{(G)}\} \sim \pi_{\theta_{\text{old}}}(\cdot|p)$
        \STATE Compute $r^{(g)} = R_{\text{ACT}}(s,y^{(g)})$ for $g=1,\ldots,G$
        \STATE Compute $\hat{A}^{(g)}$ via \cref{eq:grpo_advantage} for $g=1,\ldots,G$
    \ENDFOR
    \STATE Update $\theta$ by minimizing $\mathcal{L}_{\text{GRPO}}(\theta)$
\ENDFOR
\STATE \textbf{return} $\pi_{\theta}$
\end{algorithmic}
\end{algorithm}

\subsection{Implementation details}

We use OpenRLHF~\citep{hu2024openrlhf} for GRPO training with DeepSpeed ZeRO-3~\citep{rajbhandari2020zero} for memory efficiency. Training uses 16 NVIDIA GH200 GPUs. \Cref{tab:hyperparams} lists the Qwen3 and Olmo 3 training hyperparameters.

\begin{table}[h]
\centering
\caption{Training hyperparameters for Qwen3 and Olmo-3-7B-Instruct.}
\label{tab:hyperparams}
\begin{tabular}{ll}
\toprule
\textbf{Hyperparameter} & \textbf{Value} \\
\midrule
Base models & Qwen3-4B, Qwen3-8B, Olmo-3-7B-Instruct \\
Learning rate & $2 \times 10^{-6}$ \\
Learning-rate scheduler & Cosine \\
Warmup ratio & 0.1 \\
Batch size & 64 \\
Group size ($G$) & 16 (Qwen3-4B); 8 (Qwen3-8B and Olmo 3) \\
Candidate samples ($K$) & 1 \\
Epochs per stage (ACT / IL / RL) & 3 \\
Prompt max length & 4,096 tokens \\
Generation max length & 4,096 tokens \\
Temperature & 1.0 \\
Top-p & 0.95 \\
KL coefficient ($\beta$) & 0.0 \\
Optimizer & AdamW with offload \\
Precision & bfloat16 (BF16) \\
\bottomrule
\end{tabular}
\end{table}

\subsection{Data statistics}

\Cref{tab:datasets} summarizes the expert-data statistics across all benchmarks. Each sample corresponds to a single expert state--action pair extracted from a successful trajectory. ACT and IL methods use the same pairs to ensure a fair comparison; RL starts from the corresponding IL checkpoint. Each ACT, IL, and RL stage uses three epochs; IL$\to$IL and IL$\to$RL$\to$RL add three epochs to their respective baseline checkpoints (\cref{sec:training_controls}). For ScienceWorld, we randomly sample 10,240 state-action pairs for training from the full set of expert trajectories.

\begin{table}[h]
\centering
\caption{Dataset statistics for all training. Train samples are state-action pairs. \textit{ID}: In-Distribution, \textit{OOD}: Out-of-Distribution.}
\label{tab:datasets}
\begin{tabular}{lcccc}
\toprule
\textbf{Benchmark} & \textbf{Domain} & \textbf{Train Pairs} & \textbf{Task Types} & \textbf{Test Samples} \\
\midrule
ALFWorld      & Embodied & 10,240 & 6 & 140 (ID) / 134 (OOD) episodes \\
WebShop       & Web      & 3,071  & N/A & 500 episodes \\
ScienceWorld  & Science  & 10,240 & 30 & 149 episodes \\
\bottomrule
\end{tabular}
\end{table}

\subsection{Choice of \texorpdfstring{$K$}{K}}
\label{app:pair_quality}

We set $K=1$ in all experiments (\cref{tab:hyperparams}). Since the expert trajectory datasets already provide a sufficient number of training samples (\cref{tab:datasets}), a single alternative per state yields enough contrastive pairs for effective training without the need for $K>1$. In settings where the expert dataset is smaller, increasing $K$ can be used to produce additional contrastive pairs per state to compensate.

\subsection{Expert trajectory collection}
\label{app:expert_trajectories}

For ALFWorld, expert trajectories are collected by running the model released by \citet{fenggroup} on the ALFWorld training set. For WebShop, expert trajectories come from the official human demonstration data released with the benchmark. For ScienceWorld, expert trajectories come from the official gold trajectories released with the benchmark.

\subsection{Additional evaluation protocol}
\label{app:revision_protocol}

\paragraph{ALFWorld.} We evaluate each checkpoint on 140 unique valid-seen episodes and 134 unique valid-unseen episodes. The two splits are reported as ID and OOD success rates, respectively. Each episode runs for at most 50 interaction steps.

\paragraph{WebShop.} We evaluate each checkpoint on 500 episodes and report the task success rate.

\paragraph{ScienceWorld.} We use the official test variations with at most five variations per task, producing 149 interactive episodes. Each episode runs for at most 50 steps. The primary metric is the mean of the highest environment score reached in each episode, on a 0--100 scale, which retains credit for completed subgoals. We additionally record full task completion when the score reaches 100.

\paragraph{Unified CoT protocol.} For the prompting baselines, both Qwen3 and Olmo 3 receive the same prompt-level instruction: ``Think step by step about which action best advances the task, then give your choice.'' The CoT and no-CoT prompts differ only by this instruction.

\paragraph{Seeded evaluation.} We sample with temperature 0.6 and top-$p$ 0.95 and repeat each evaluation with three runs. All reported values with error bars are the mean $\pm$ standard deviation across these runs.

\subsection{Baseline and control implementations}
\label{app:baseline_details}

\paragraph{Reflection baselines.}
\begin{itemize}
    \item \textbf{Early Experience}: following the self-reflection variant of \citet{zhang2025agent}, executes both expert and alternative actions in the environment, uses the resulting next-state differences to synthesize reflection targets, and trains the model through next-token prediction to reproduce each reflection together with the expert action. The resulting data is mixed with the standard expert-action data.
    \item \textbf{SAND}: samples $N=5$ candidate actions at each expert decision point, executes them in the environment to obtain feedback, synthesizes an action-deliberation target from execution-grounded critiques, and applies SFT to the deliberation followed by the expert action~\citep{xia2025sand}.
    \item \textbf{Agent-R}: rolls out the current policy, identifies its first deviation from the expert trajectory, and constructs revision trajectories by combining the erroneous prefix, a reflection, the expert correction, and the remaining expert suffix; these trajectories are mixed with expert trajectories for SFT~\citep{yuan2025agent}.
\end{itemize}
All three baselines optimize next-token cross-entropy rather than a policy-gradient objective.

\paragraph{Same-action-pair controls.}
Each control uses ACT's expert--alternative pairs, changes only the stage-1 objective, and then follows the same IL$\to$RL downstream pipeline:
\begin{itemize}
    \item \textbf{Full pipeline (SFT-on-choice)}: replaces the ACT-stage RL objective with supervised fine-tuning on the correct action only, without training a rationale.
    \item \textbf{Full pipeline (DPO)}: replaces the ACT-stage RL objective with DPO, treating the expert action as chosen and the alternative as rejected.
    \item \textbf{Full pipeline (no-reasoning)}: keeps the ACT-stage RL objective and reward but removes the reasoning instruction, requiring the model to select an action directly.
\end{itemize}

\subsection{Causal rationale intervention protocol}
\label{app:rationale_protocol}

We sample 400 held-out ALFWorld action pairs and compare the original Qwen3-8B checkpoint with the standalone ACT-stage checkpoint under three inference conditions. In the \emph{free} condition, the model generates its reasoning and choice normally. In the \emph{no-rationale} condition, the reasoning block is pre-closed and the model directly selects an action. In the \emph{incorrect-rationale} condition, we inject a plausible rationale favoring the alternative action as if it had already been generated by the model, then ask the model to complete only the final choice. We report choice accuracy and the fraction of examples whose final choice follows the injected alternative. The intervention changes only the rationale context; the state, candidate actions, checkpoint, and decoding protocol remain fixed~\citep{han2026zerotuning}.

\begin{table}[h]
\centering
\caption{Exact results for the causal rationale intervention in \cref{fig:rationale_intervention} (\%).}
\label{tab:rationale_intervention}
\begin{tabular}{lcc}
\toprule
\textbf{Condition} & \textbf{Original Qwen3-8B} & \textbf{ACT checkpoint} \\
\midrule
Free choice accuracy & 45.00 & \textbf{86.50} \\
No-rationale accuracy & 46.25 & \textbf{80.75} \\
Incorrect-rationale accuracy & 2.25 & \textbf{50.75} \\
Follows injected action $\downarrow$ & 81.75 & \textbf{28.25} \\
\bottomrule
\end{tabular}
\end{table}

\subsection{Cross-size transfer to Qwen3-4B}
\label{sec:cross_model}

ACT requires sampling alternative actions to construct its contrastive pairs. We test whether pairs collected from a larger model can be reused by training Qwen3-4B on ALFWorld with ACT pairs collected entirely from Qwen3-8B, without re-collection or adaptation. The 4B experiments use the same method definitions as the Qwen3-8B main experiments: we compare IL with ACT$\to$IL, and IL$\to$RL w/o ACT with the full ACT pipeline, ACT$\to$IL$\to$RL.

\begin{table}[h]
\centering
\caption{Cross-size transfer results on Qwen3-4B for ALFWorld. Values are ID and OOD success rates (\%). All ACT pairs used by the ACT-based methods were collected from Qwen3-8B.}
\label{tab:cross_model}
\begin{tabular}{lcc}
\toprule
\textbf{Method} & \textbf{ID} & \textbf{OOD} \\
\midrule
Prompt w/o CoT thinking & 13.57 & 8.96 \\
Prompt w/ CoT thinking & 50.71 & 29.85 \\
\midrule
ACT stage & 71.43 & 62.69 \\
\midrule
Imitation Learning (IL) & 85.00 & 83.58 \\
Early Experience & 88.57 & 88.06 \\
ACT $\to$ IL & 88.57 & 91.04 \\
\midrule
IL $\to$ RL w/o ACT & 91.43 & 88.81 \\
Full ACT pipeline & \textbf{92.14} & \textbf{91.79} \\
\bottomrule
\end{tabular}
\end{table}

As shown in \cref{tab:cross_model}, transferred ACT data improves both downstream pipelines on Qwen3-4B in both ID and OOD settings. Thus, ACT pairs collected from Qwen3-8B remain effective for a smaller model without re-collection or adaptation.

\newpage
\subsection{Full general-reasoning results}
\label{app:plotted_results}

Table~\ref{tab:general_reasoning} reports the full results on MATH-500 and GPQA-Diamond, with accuracy (\%) expressed as the mean $\pm$ standard deviation over three runs.

\begin{table}[h]
\centering
\caption{Full general-reasoning results corresponding to \cref{fig:general_reasoning}. ACT denotes the standalone ACT-stage checkpoint, without downstream IL or RL.}
\label{tab:general_reasoning}
\begin{tabular}{lcc}
\toprule
\textbf{Method} & \textbf{MATH-500} & \textbf{GPQA-Diamond} \\
\midrule
Prompt w/o CoT thinking & 78.60$\pm$0.33 & 42.93$\pm$1.09 \\
Prompt w/ CoT thinking & 86.93$\pm$0.74 & 51.52$\pm$1.89 \\
\midrule
Imitation Learning & 87.00$\pm$0.33 & 44.61$\pm$0.95 \\
Early Experience (Self-Reflection) & 86.86$\pm$0.25 & 51.85$\pm$0.63 \\
SAND & 81.00$\pm$0.86 & 52.53$\pm$1.65 \\
Agent-R & 86.53$\pm$0.81 & 52.86$\pm$1.04 \\
IL $\to$ RL w/o ACT & 87.07$\pm$0.77 & 52.36$\pm$1.32 \\
Standalone ACT stage & \textbf{87.73$\pm$0.19} & \textbf{53.37$\pm$0.63} \\
\bottomrule
\end{tabular}
\end{table}
\clearpage

\subsection{ACT prompt templates}
\label{app:prompt_templates}

The following templates are used during the ACT stage. In each template, \texttt{action\_1} and \texttt{action\_2} contain the expert and alternative actions in randomized order.

\paragraph{ALFWorld.}
The ACT prompt for ALFWorld is shown in \Cref{fig:prompt_critical_alfworld}.

\begin{figure}[h]
\centering
\begin{tcolorbox}[
  enhanced,
  boxrule=0.5pt,
  colback=AliceBlue,
  colframe=Navy,
  colbacktitle=Navy,
  fonttitle=\bfseries\color{white},
  title=ACT Prompt (ALFWorld),
  arc=2mm,
  attach boxed title to top left={yshift=-2.5mm, xshift=3mm},
  boxed title style={arc=1mm},
  boxsep=5pt,
  left=5pt,
  right=5pt,
  top=4mm,
]
\small\texttt{%
You are an expert agent operating in the ALFWorld Environment.\\[2ex]
Task: \{task\_description\}\\[2ex]
Prior to this step, you have already taken \{n\} step(s).\\
Below are the most recent \{k\} observations and actions:\\
\{history\}\\[2ex]
You are now at step \{current\_step\} and your current observation is:\\
\{observation\}\\[2ex]
Your admissible actions of the current situation are:\\
\{admissible\_actions\}\\[2ex]
You need to decide which action is better for completing the task.\\[2ex]
Output format requirements:\\
1. First, think about which action is better and why.\\
2. Then, output the chosen action content inside <action>...</action> tags.\\
3. You must output the actual action text, NOT `Action 1' or `Action 2'.\\[2ex]
Candidate Actions for the NEXT step:\\
Action 1: \{action\_1\}\\
Action 2: \{action\_2\}\\[2ex]
Which action is better as the NEXT step? Think about why, then output the chosen action inside <action>...</action> tags.
}
\end{tcolorbox}
\caption{The ACT prompt for ALFWorld. The model is presented with the full context followed by two candidate actions and is asked to select the better one with reasoning.}
\label{fig:prompt_critical_alfworld}
\end{figure}

\clearpage

\paragraph{WebShop.}
The ACT prompt for WebShop is shown in \Cref{fig:prompt_critical_webshop}.

\begin{figure}[h]
\centering
\begin{tcolorbox}[
  enhanced,
  boxrule=0.5pt,
  colback=AliceBlue,
  colframe=Navy,
  colbacktitle=Navy,
  fonttitle=\bfseries\color{white},
  title=ACT Prompt (WebShop),
  arc=2mm,
  attach boxed title to top left={yshift=-2.5mm, xshift=3mm},
  boxed title style={arc=1mm},
  boxsep=5pt,
  left=5pt,
  right=5pt,
  top=4mm,
]
\small\texttt{%
You are an expert shopping assistant navigating an e-commerce website.\\[2ex]
Task: \{task\_description\}\\[2ex]
Prior to this step, you have already taken \{n\} step(s).\\
Below are the most recent \{k\} observations and actions:\\
\{history\}\\[2ex]
You are now at step \{current\_step\} and your current observation is:\\
\{observation\}\\[2ex]
Your admissible actions of the current situation are:\\
\{admissible\_actions\}\\[2ex]
You need to decide which action is better for completing the task.\\[2ex]
Output format requirements:\\
1. First, think about which action is better and why.\\
2. Then, output the chosen action content inside <action>...</action> tags.\\
3. You must output the actual action text, NOT `Action 1' or `Action 2'.\\[2ex]
Candidate Actions for the NEXT step:\\
Action 1: \{action\_1\}\\
Action 2: \{action\_2\}\\[2ex]
Which action is better as the NEXT step? Think about why, then output the chosen action inside <action>...</action> tags.
}
\end{tcolorbox}
\caption{The ACT prompt for WebShop. The model is presented with the full context followed by two candidate actions and is asked to select the better one with reasoning.}
\label{fig:prompt_critical_webshop}
\end{figure}

\clearpage

\paragraph{ScienceWorld.}
The ACT prompt for ScienceWorld is shown in \Cref{fig:prompt_critical_scienceworld}.

\begin{figure}[h]
\centering
\begin{tcolorbox}[
  enhanced,
  boxrule=0.5pt,
  colback=AliceBlue,
  colframe=Navy,
  colbacktitle=Navy,
  fonttitle=\bfseries\color{white},
  title=ACT Prompt (ScienceWorld),
  arc=2mm,
  attach boxed title to top left={yshift=-2.5mm, xshift=3mm},
  boxed title style={arc=1mm},
  boxsep=5pt,
  left=5pt,
  right=5pt,
  top=4mm,
]
\small\texttt{%
You are an expert science tutor guiding a student through hands-on experiments in the ScienceWorld environment.\\[2ex]
Task: \{task\_description\}\\[2ex]
Prior to this step, you have already taken \{n\} step(s).\\
Below are the most recent \{k\} observations and actions:\\
\{history\}\\[2ex]
You are now at step \{current\_step\} and your current observation is:\\
\{observation\}\\[2ex]
Your admissible actions of the current situation are:\\
\{admissible\_actions\}\\[2ex]
You need to decide which action is better for completing the task.\\[2ex]
Output format requirements:\\
1. First, think about which action is better and why.\\
2. Then, output the chosen action content inside <action>...</action> tags.\\
3. You must output the actual action text, NOT `Action 1' or `Action 2'.\\[2ex]
Candidate Actions for the NEXT step:\\
Action 1: \{action\_1\}\\
Action 2: \{action\_2\}\\[2ex]
Which action is better as the NEXT step? Think about why, then output the chosen action inside <action>...</action> tags.
}
\end{tcolorbox}
\caption{The ACT prompt for ScienceWorld. The model is presented with the full context followed by two candidate actions and is asked to select the better one with reasoning.}
\label{fig:prompt_critical_scienceworld}
\end{figure}

\clearpage

%% file: sections/a1_agentic_case.tex
\section{Additional case studies: agentic task performance}
\label{app:agentic_case_studies}

We provide case studies illustrating failure recovery by the full ACT pipeline and characteristic decision-making failures of IL.

\subsection{Failure recovery through self-critique}
\label{app:failure_recovery}

In ALFWorld, the environment returns ``Nothing happens.'' when an action fails. In the example below, the IL model repeats the same failed action until termination. In contrast, the model trained with the full ACT pipeline diagnoses the root cause through its internal reasoning and issues the correct corrective action. This behavior is consistent with the action-evaluation capability introduced by ACT. \Cref{fig:case_alfworld} shows the two representative traces.

\begin{figure}[ht]
\centering
\begin{minipage}[t]{0.48\textwidth}
\begin{tcolorbox}[
  enhanced,
  boxrule=0.5pt,
  colback=AliceBlue,
  colframe=Navy,
  colbacktitle=Navy,
  fonttitle=\bfseries\color{white},
  title=IL Task,
  arc=2mm,
  attach boxed title to top left={yshift=-2.5mm, xshift=3mm},
  boxed title style={arc=1mm},
  boxsep=5pt,
  left=5pt,
  right=5pt,
  top=4mm,
  equal height group=toprow,
]
\small
Step 7: take cloth 1 from countertop 1\\
Step 8: clean cloth 1 with sinkbasin 1\\
Step 9: put cloth 1 in/on cabinet 1\\
\quad\small\itshape $\leftarrow$ fails\\[1mm]
\normalfont\small
Step 10: put cloth 1 in/on cabinet 1\\
Step 11: put cloth 1 in/on cabinet 1\\
Step 12: put cloth 1 in/on cabinet 1\\
\itshape\quad\vdots\\
(Repeats for 30+ steps until termination)
\end{tcolorbox}
\vspace{2mm}
\begin{tcolorbox}[
  enhanced,
  boxrule=0.5pt,
  colback=LavenderBlush,
  colframe=Crimson,
  colbacktitle=Crimson,
  fonttitle=\bfseries\color{white},
  title=IL Result: Failed $\times$,
  arc=2mm,
  attach boxed title to top left={yshift=-2.5mm, xshift=3mm},
  boxed title style={arc=1mm},
  boxsep=5pt,
  left=5pt,
  right=5pt,
  top=4mm,
  equal height group=bottomrow,
]
\small
The IL model repeats the identical failed action for over 30 steps until termination. Having never seen failure states during imitation learning, it has no mechanism to diagnose the error or propose an alternative.
\end{tcolorbox}
\end{minipage}
\hfill
\begin{minipage}[t]{0.48\textwidth}
\begin{tcolorbox}[
  enhanced,
  boxrule=0.5pt,
  colback=AliceBlue,
  colframe=Navy,
  colbacktitle=Navy,
  fonttitle=\bfseries\color{white},
  title=Full ACT Pipeline Task,
  arc=2mm,
  attach boxed title to top left={yshift=-2.5mm, xshift=3mm},
  boxed title style={arc=1mm},
  boxsep=5pt,
  left=5pt,
  right=5pt,
  top=4mm,
  equal height group=toprow,
]
\small
Step 4: clean spoon 2 with sinkbasin 1\\
Step 5: use spoon 2 on diningtable 1\\
\quad\small\itshape $\leftarrow$ ``Nothing happens.''\\[1mm]
\normalfont\small
Step 6: put spoon 2 in/on diningtable 1\\
\quad\small\itshape $\leftarrow$ ``Nothing happens.''\\[1mm]
\normalfont\small
Step 7: put spoon 2 in/on diningtable 1\\
\quad\small\itshape $\leftarrow$ ``Nothing happens.''
\end{tcolorbox}
\vspace{2mm}
\begin{tcolorbox}[
  enhanced,
  boxrule=0.5pt,
  colback=Honeydew,
  colframe=ForestGreen,
  colbacktitle=ForestGreen,
  fonttitle=\scriptsize\bfseries\color{white},
  title=Full ACT Pipeline: Self-Critique \& Recovery (\checkmark),
  arc=2mm,
  attach boxed title to top left={yshift=-2.5mm, xshift=3mm},
  boxed title style={arc=1mm},
  boxsep=5pt,
  left=5pt,
  right=5pt,
  top=4mm,
  equal height group=bottomrow,
]
\small\itshape
I cleaned spoon 2 with sinkbasin 1. Then I tried to put spoon 2 in/on diningtable 1, but nothing happened. I did this twice. \textbf{Wait\ldots{} the task is to put it IN the dining table. I am currently at sinkbasin 1. I need to GO TO the dining table first before I can put it there!}

\vspace{2mm}
\normalfont\small\textbf{Action: go to diningtable 1} \checkmark
\end{tcolorbox}
\end{minipage}

\caption{Failure recovery on ALFWorld. \textbf{Left:} The IL model enters an infinite loop, repeating a failed action for over 30 steps until termination. \textbf{Right:} The model trained with the full ACT pipeline encounters the same type of failure but uses its internal reasoning to diagnose the root cause (wrong location), break the loop, and issue the correct navigation command.}
\label{fig:case_alfworld}
\end{figure}

\subsection{Rigid execution without state awareness}
\label{app:rigid_execution}

IL trains agents to reproduce demonstrated action patterns. When the environment state deviates from those patterns, an IL model may fail to detect the mismatch and adjust. \Cref{fig:case_webshop} illustrates this on a WebShop task: the user requests men's shirts with specific attributes \emph{priced below \$50}. The IL model follows a rigid sequence (search $\to$ click item $\to$ select attributes $\to$ buy) without checking whether the item satisfies the constraints. It clicks ``Buy Now'' on a \$55 item, violating the price requirement and receiving a score of 0.

This example highlights the type of state-aware action evaluation targeted by ACT: before proceeding, the model should assess whether the current product page satisfies the task constraints and search again when it does not.

\begin{figure}[ht]
\centering
\begin{tcolorbox}[
  enhanced,
  boxrule=0.5pt,
  colback=AliceBlue,
  colframe=Navy,
  colbacktitle=Navy,
  fonttitle=\bfseries\color{white},
  title=WebShop Task,
  arc=2mm,
  attach boxed title to top left={yshift=-2.5mm, xshift=3mm},
  boxed title style={arc=1mm},
  boxsep=5pt,
  left=5pt,
  right=5pt,
  top=4mm,
]
\small
Find me wash cold, machine wash men's shirts with polyester heathers, heathers cotton\ldots{} and \textbf{price lower than 50.00 dollars}.
\end{tcolorbox}

\vspace{2mm}

\begin{tcolorbox}[
  enhanced,
  boxrule=0.5pt,
  colback=LavenderBlush,
  colframe=Crimson,
  colbacktitle=Crimson,
  fonttitle=\bfseries\color{white},
  title=IL Trajectory (Score: 0.0 $\times$),
  arc=2mm,
  attach boxed title to top left={yshift=-2.5mm, xshift=3mm},
  boxed title style={arc=1mm},
  boxsep=5pt,
  left=5pt,
  right=5pt,
  top=4mm,
]
\small
\begin{tabular}{@{}ll@{}}
Step 1: & search[men's shirts polyester heathers] \\
Step 2: & click[B07XJD\ldots] \\
        & \quad\small\itshape Observation: \textbf{Price: \$55.00} (exceeds \$50 budget) \\
Step 3: & click[black] \\
Step 4: & click[small] \\
Step 5: & click[Buy Now] \quad\small $\leftarrow$ \textit{buys item despite constraint violation}
\end{tabular}
\end{tcolorbox}

\caption{Rigid execution on WebShop. The IL model follows a scripted sequence (search $\to$ click $\to$ buy) without checking whether the item satisfies the price constraint (\$55 $>$ \$50 budget), resulting in a failed purchase.}
\label{fig:case_webshop}
\end{figure}

%% file: sections/a2_general_case.tex
\section{Additional case studies: general reasoning after agentic training}
\label{app:case_studies}

We provide case studies analyzing the standalone ACT-stage checkpoint's self-verification behavior and reasoning degradation after IL, based on Qwen3-8B traces from GPQA-Diamond.

\subsection{Self-verification behavior}
\label{app:self_verification}

On certain difficult problems, the standalone ACT-stage checkpoint exhibits \emph{self-verification} behavior: after performing an initial derivation, the model checks its answer by substituting back into the original equations. \Cref{fig:case_verification} shows one such example on a particle physics problem: after deriving the kinetic energies, the standalone ACT-stage checkpoint substitutes each answer option back into the energy conservation equation to verify consistency, systematically eliminating incorrect options. The corresponding original-model trace performs the initial derivation but does not systematically verify the options.

\begin{figure}[ht]
\centering
\begin{tcolorbox}[
  enhanced,
  boxrule=0.5pt,
  colback=AliceBlue,
  colframe=Navy,
  colbacktitle=Navy,
  fonttitle=\bfseries\color{white},
  title=GPQA-Diamond \#12: Pion Decay Kinetic Energy,
  arc=2mm,
  attach boxed title to top left={yshift=-2.5mm, xshift=3mm},
  boxed title style={arc=1mm},
  boxsep=5pt,
  left=5pt,
  right=5pt,
  top=4mm,
]
\small
Find the kinetic energy of product particles in $\pi^+ \to \mu^+ + \nu$, where $\pi^+$ is stationary. Rest mass of $\pi^+$ and $\mu^+$ is 139.6 MeV and 105.7 MeV respectively.

Options: (A) $KE_\mu = 4.12$ MeV, $KE_\nu = 29.8$ MeV \quad (B) $KE_\mu = 7.2$ MeV, $KE_\nu = 32.8$ MeV \quad (C) $KE_\mu = 3.52$ MeV, $KE_\nu = 20.8$ MeV \quad (D) $KE_\mu = 2.84$ MeV, $KE_\nu = 26.8$ MeV
\end{tcolorbox}

\vspace{2mm}

\begin{tcolorbox}[
  enhanced,
  boxrule=0.5pt,
  colback=Honeydew,
  colframe=ForestGreen,
  colbacktitle=ForestGreen,
  fonttitle=\bfseries\color{white},
  title=Standalone ACT-Stage Trace: Self-Verification (Correct \checkmark),
  arc=2mm,
  attach boxed title to top left={yshift=-2.5mm, xshift=3mm},
  boxed title style={arc=1mm},
  boxsep=5pt,
  left=5pt,
  right=5pt,
  top=4mm,
]
\small\itshape
[\ldots derives kinetic energies via conservation of energy and momentum\ldots]

\normalfont\small\textbf{Verification of each option against energy conservation ($E_\text{total} = 139.6$ MeV):}

\small\itshape
Option C: $KE_\mu = 3.52$ MeV, $KE_\nu = 20.8$ MeV. Then, total energy would be $105.7 + 3.52 = 109.22$ MeV for the muon. Neutrino energy is 20.8 MeV. Total energy: $109.22 + 20.8 = 130.02$ MeV. \textbf{Which is less than 139.6. So that's not possible.}

Option D: $KE_\mu = 2.84$ MeV, $KE_\nu = 26.8$ MeV. Total energy: $105.7 + 2.84 = 108.54$ MeV for muon. Neutrino energy is 26.8 MeV. Total energy is $108.54 + 26.8 = 135.34$ MeV. \textbf{Still less than 139.6.}

So, only option A gives the total energy matching. Therefore, the correct answer is A.

\vspace{2mm}
\normalfont\small\textbf{Answer: A} \checkmark
\end{tcolorbox}

\caption{Self-verification behavior observed in the standalone ACT-stage checkpoint on GPQA-Diamond. After deriving the kinetic energies, the checkpoint substitutes each answer option back into the energy conservation equation, eliminating inconsistent options.}
\label{fig:case_verification}
\end{figure}

\subsection{Reasoning collapse after IL}
\label{app:reasoning_collapse}

IL on agentic data fine-tunes the model on short, action-heavy expert trajectories that contain minimal extended reasoning. By imitating these behavioral patterns, the model can lose some of its general reasoning capability: the chain-of-thought behavior that the original model possesses may be disrupted by the short-sequence, action-centric distribution of agentic data. We refer to this observed degradation as \emph{reasoning collapse}; one representative trace is shown below.

\paragraph{Unfocused meandering.} Even when the IL model produces a reasoning trace, its reasoning can become less focused. \Cref{fig:case_collapse_meander} shows a high-energy physics problem (GPQA-Diamond \#7) about gamma-ray annihilation with CMB photons. The standalone ACT-stage checkpoint produces a focused 10{,}669-character derivation that sets up the threshold-energy condition and arrives at the correct answer. The IL model generates a 37{,}924-character trace, 3.6$\times$ longer, but wanders through vague recollections and contradictory estimates before guessing incorrectly.

\begin{figure}[ht]
\centering
\begin{tcolorbox}[
  enhanced,
  boxrule=0.5pt,
  colback=AliceBlue,
  colframe=Navy,
  colbacktitle=Navy,
  fonttitle=\bfseries\color{white},
  title=GPQA-Diamond \#7: Gamma-Ray CMB Annihilation Threshold,
  arc=2mm,
  attach boxed title to top left={yshift=-2.5mm, xshift=3mm},
  boxed title style={arc=1mm},
  boxsep=5pt,
  left=5pt,
  right=5pt,
  top=4mm,
]
\small
The universe is filled with the Cosmic Microwave Background. Consider the annihilation of high-energy $\gamma$-rays with a CMB photon into electron-positron pairs, i.e., $\gamma\gamma\rightarrow e^+e^-$. Find the energy of $\gamma$-rays whose lifetimes are limited by this process. The average energy of a CMB photon is $10^{-3}$ eV.
\end{tcolorbox}

\vspace{2mm}

\begin{minipage}[t]{0.48\textwidth}
\begin{tcolorbox}[
  enhanced,
  boxrule=0.5pt,
  colback=Honeydew,
  colframe=ForestGreen,
  colbacktitle=ForestGreen,
  fonttitle=\bfseries\color{white},
  title=Standalone ACT-Stage Trace (Correct \checkmark),
  arc=2mm,
  attach boxed title to top left={yshift=-2.5mm, xshift=3mm},
  boxed title style={arc=1mm},
  boxsep=5pt,
  left=5pt,
  right=5pt,
  top=4mm,
]
\small\itshape
For $\gamma\gamma \to e^+e^-$, the total energy must be at least $2m_e c^2$. For a head-on collision, the threshold condition is $E_\gamma \cdot E_\text{CMB} \geq (m_e c^2)^2$. [\ldots] So $E_\gamma \geq (0.511 \text{ MeV})^2 / (10^{-3} \text{ eV}) = 2.61 \times 10^{14} \text{ eV} \approx 2.6 \times 10^5$ GeV. The gamma rays with energy just above the threshold would have their lifetimes limited by this process. Therefore, the answer is the threshold energy.

\vspace{2mm}
\normalfont\small\textbf{Answer: C ($2.6 \times 10^5$ GeV)} \checkmark \quad (10{,}669 chars)
\end{tcolorbox}
\end{minipage}
\hfill
\begin{minipage}[t]{0.48\textwidth}
\begin{tcolorbox}[
  enhanced,
  boxrule=0.5pt,
  colback=LavenderBlush,
  colframe=Crimson,
  colbacktitle=Crimson,
  fonttitle=\bfseries\color{white},
  title=IL Reasoning Trace (Wrong $\times$),
  arc=2mm,
  attach boxed title to top left={yshift=-2.5mm, xshift=3mm},
  boxed title style={arc=1mm},
  boxsep=5pt,
  left=5pt,
  right=5pt,
  top=4mm,
]
\small
[\ldots after 37{,}924 characters of unfocused derivation\ldots]

\itshape
However, I recall that the energy at which gamma rays can interact with CMB photons to produce $e^+e^-$ pairs is approximately $10^4$ GeV. [\ldots] Given that, the answer is likely option B or C. But I'm not certain. [\ldots]

Given the time I've spent and the lack of progress, I'll go with option B: $1.8 \times 10^5$ GeV.

\vspace{2mm}
\normalfont\small\textbf{Answer: B ($1.8 \times 10^5$ GeV)} $\times$ \quad (37{,}924 chars)
\end{tcolorbox}
\end{minipage}

\caption{Reasoning collapse after IL: unfocused meandering. On a high-energy physics threshold problem, the standalone ACT-stage checkpoint produces a focused derivation (10K chars), while IL generates 3.6$\times$ as much text (38K chars) yet wanders through contradictory estimates and guesses incorrectly.}
\label{fig:case_collapse_meander}
\end{figure}

This example is consistent with ACT preserving more of the original model's reasoning behavior than IL: ACT optimizes correct action selection via RL rather than imitating short, action-centric targets. We treat the trace as qualitative evidence rather than a general claim about every problem instance.